%% file: main.tex
\pdfoutput=1

\documentclass[11pt]{article}

\usepackage{latex/acl}
\usepackage{times}
\usepackage{enumitem}
\usepackage{multirow}
\usepackage{multicol}
\usepackage{float}
\usepackage{latexsym}

\usepackage{booktabs} 
\usepackage[normalem]{ulem}
\usepackage{xcolor} 
\usepackage{algorithm}
\usepackage[noend]{algpseudocode}
\usepackage{amsmath}
\usepackage{mathrsfs}
 \usepackage{amssymb}
\usepackage{subfigure}
\usepackage{makecell}
\usepackage{mathtools}
\usepackage[font=rm]{caption}
\DeclareCaptionType{copyrightbox}
\usepackage{shortvrb}
\usepackage{tabularx}
\usepackage{verbatim}
\usepackage{xspace}
\usepackage{listings}
\usepackage{tikz}
\usepackage{verbatim}
\usetikzlibrary{trees}

\usepackage{forest}
\forestset{
    nice empty nodes/.style={
        for tree={
            s sep=0.1em, 
            l sep=0.33em,
            inner ysep=0.4em, 
            inner xsep=0.05em,
            l=0,
            calign=midpoint,
            fit=tight,
            where n children=0{
               tier=word,
               minimum height=1.25em,
            }{},
            where n children=2{
               l-=1em,
            }{},
            parent anchor=south,
            child anchor=north,
            delay={if content={}{
                    inner sep=0pt,
                    edge path={\noexpand\path [\forestoption{edge}] 
                    			(!u.parent anchor) 
                               -- (.south)\forestoption{edge label};}
                }{}}
        },
    },
}
\usepackage{fontawesome}
\usepackage[multiple]{footmisc}
\usepackage[all]{nowidow}
\usepackage{balance}
\usepackage{microtype}
\DeclareMathOperator{\MLP}{MLP}

\DeclareMathOperator{\Softmax}{softmax}

\DeclareMathOperator{\for}{for}

\newcommand{\gen}{\textsc{Gen}\xspace}
\newcommand{\comp}{\textsc{Comp}\xspace}
\usepackage{inconsolata}

\title{Generative Pretrained Structured Transformers:\\
Unsupervised Syntactic Language Models at Scale}

\author{Xiang Hu\footnotemark[3]~\thanks{~~Equal contribution, see appendix~\ref{appdx:contribution_statement} for details.} \quad Pengyu Ji\footnotemark[4] \footnotemark[1] \quad Qingyang Zhu\footnotemark[4] \quad Wei Wu\footnotemark[3] \thanks{~Corresponding authors.}\quad Kewei Tu\footnotemark[4] \footnotemark[2]\\
 Ant Group\footnotemark[3] \\
 \tt \{aaron.hx, congyue.ww\}@antgroup.com\footnotemark[3] \\
ShanghaiTech University\footnotemark[4] \\
 \tt \{jipy2023, zhuqy, tukw\}@shanghaitech.edu.cn\footnotemark[4]  \\}

\begin{document}
\maketitle
\begin{abstract}
A syntactic language model (SLM) incrementally generates a sentence with its syntactic tree in a left-to-right manner.
We present \textbf{G}enerative \textbf{P}retrained \textbf{S}tructured \textbf{T}ransformers (GPST), an unsupervised SLM at scale capable of being pre-trained from scratch on raw texts with high parallelism. 
GPST circumvents the limitations of previous SLMs such as relying on gold trees and sequential training.
It consists of two components, a usual SLM supervised by a uni-directional language modeling loss, and an additional composition model, which induces syntactic parse trees and computes constituent representations, supervised by a bi-directional language modeling loss. We propose a representation surrogate to enable joint parallel training of the two models in a hard-EM fashion.
We pre-train GPST on OpenWebText, a corpus with $9$ billion tokens, and demonstrate the superiority of GPST over GPT-2 with a comparable size in numerous tasks covering both language understanding and language generation. Meanwhile, GPST also significantly outperforms existing unsupervised SLMs on left-to-right grammar induction, while holding a substantial acceleration on training.\footnote{~Code will be released at \url{https://github.com/ant-research/StructuredLM_RTDT}.}

\end{abstract}

\input{intro.tex}
\input{related_works.tex}
\input{alg.tex}
\input{experiments.tex}
\input{conclusion.tex}
\input{limitation.tex}

\clearpage
\bibliography{acl2024}

\clearpage
\appendix
\section{Appendix}
\input{appendix.tex}
\end{document}

%% file: intro.tex
\section{Introduction}
Pre-training a Transformer architecture~\cite{DBLP:conf/nips/VaswaniSPUJGKP17} as a large language model has dominated the field of natural language processing (NLP) \cite{DBLP:conf/naacl/DevlinCLT19,liu2019roberta,radford2018improving,radfordlanguage,brown2020language,ouyang2022training}. While Transformer language models have exhibited remarkable performance over various downstream NLP tasks~\cite{bang2023multitask},
the recursive compositions behind language are represented in an implicit and entangled form.
In contrast, human language understanding exhibits explicit composition decisions, as exemplified by the garden path sentence~\cite{Dynel2009HumorousGA}~``Time flies like an arrow; Fruit flies like a banana'',
where distinct syntactic configurations yield vastly divergent meanings\footnote{(Fruit (flies (like a banana))) or ((Fruit flies) (like (a banana)))}. 
In addition, human infants acquire such compositional capability without supervision~\cite{581ed463-666b-3bf7-9873-eb3cc912964d}. These phenomena motivate us to explore an \emph{unsupervised} approach to learning \emph{explicit} compositions in language modeling.

A typical approach to achieve language modeling with explicit composition is to model the joint distribution of words 
and a syntactic tree 
within the framework of syntactic language models (SLMs)~\cite{dyer-etal-2016-recurrent}. Though there has been a long line of research on SLMs, they are rarely exploited as the backbone in state-of-the-art language modeling due to poor scalability. Recent Transformer-based SLMs~\cite{sartran2022transformer,murty2023pushdown} require annotated parse trees or supervised parsers as structural supervision, leading to limited training data scales and domain adaption issues~\cite{DBLP:conf/naacl/McCloskyCJ06}. 
On the other hand, in unsupervised SLMs~\cite{kim2019unsupervised,shen2019ordered}, 
non-terminal constituents are composed from their sub-constituents sequentially in a left-to-right manner, resulting in data dependencies that impede training parallelism.

In this work, we aim to pre-train an SLM \emph{at scale} on raw texts.
To this end, we propose \textbf{G}enerative \textbf{P}retrained \textbf{S}tructured \textbf{T}ransformer~(GPST), an unsupervised SLM with the Transformer architecture as a backbone.
A common practice in existing unsupervised SLMs is to learn structures by a uni-directional language modeling loss (LM loss).
However, we empirically find such an asymmetric loss with only right-to-left feedback results in branching biases in the induced parse trees.
Based on the insight, we propose two components in GPST, a composition model performing structural learning supervised by a bi-directional LM loss, and a generative model for uni-directional syntactic language modeling.
Specifically, we train the GPST in a fashion similar to hard-EM ~\cite{liang-etal-2017-neural}: in E-step, 
the composition model runs a pruned deep inside-outside encoder to induce a parse tree and compute inside and outside representations of constituents simultaneously within logarithmic steps~\cite{hu2024augmenting};
while in M-step, we update all parameters of GPST by minimizing both the bi-directional (reconstructing the sentence from outside representations) and uni-directional LM loss given the induced tree. 
The key in the M-step lies in using the inside representations of constituents computed by the composition model as a surrogate of inputs for the generative model, which enjoys two advantages. First, the representations of all constituents pre-computed in the E-step can be simultaneously fed into the generative model, which breaks the data dependencies and facilitates training parallelism. Second, with these representations participating in generation, the uni-directional LM loss in the M-step could be back-propagated to not only the generative model but the composition model used in the E-step as well.

In experiments, we pre-train GPSTs with sizes comparable to those of GPT-2$_\text{small}$ and GPT-2$_\text{medium}$ on OpenWebText~\cite{Gokaslan2019OpenWeb}($\sim$ 9 billion tokens), and evaluate the models on various tasks including language understanding, language generation, and grammar induction. 
GPST demonstrates an approximately 60-fold training acceleration and over 15\% absolute increase in left-to-right grammar induction in comparison with existing unsupervised SLMs.
Meanwhile, GPST also shows advantages over GPT-2 across almost all language understanding/generation benchmarks. GPST provides constituent-level interfaces that are not inherently possessed by the conventional Transformer-based language models, and thus exhibits great potential to enhance interpretability~\cite{DBLP:conf/iclr/HuKT23}, support multi-modality~\cite{DBLP:conf/iclr/WanHZT22}, and improve dense retrieval in the future.
Our contributions are three-fold:
\begin{itemize}[leftmargin=*,noitemsep,nolistsep]
\item We propose an SLM consisting of a composition model in addition to a generative model, which can be trained without gold trees via a novel approach akin to hard-EM.
\item We propose a representation surrogate to enable joint parallel training of all components.
\item To the best of our knowledge, GPST is the first unsupervised SLM able to be pre-trained from scratch on billions of tokens and surpass GPT-2 on various benchmarks. The experimental results demonstrate the potential of GPST as a backbone for large language models. 
\end{itemize}

%% file: related_works.tex
\section{Related work}
\paragraph{Syntactic Language Models.} There have been extensive studies on syntactic language modeling~\cite{Baker1979TrainableGF,jelinek-lafferty-1991-computation,DBLP:conf/acl/Chelba97,DBLP:journals/csl/ChelbaJ00,DBLP:conf/nips/VinyalsKKPSH15, charniak2016parsing, dyer-etal-2016-recurrent,qian2021structural,yoshida-oseki-2022-composition}, in which words and constituent symbols are mixed up and generated in a left-to-right manner. Recent works~\cite{sartran2022transformer, murty2023pushdown} utilize Transformers to parameterize action probability distributions, but relies on annotated parse trees or parsers trained on gold trees as structural guidance. Besides, unsupervised SLMs are also explored, by differentiable structured hidden layers~\cite{DBLP:conf/iclr/KimDHR17,DBLP:conf/iclr/ShenLHC18,shen2019ordered,dusell2024stack}, reinforcement learning approaches~\cite{DBLP:conf/iclr/YogatamaBDGL17}, or variational approximations~\cite{DBLP:conf/aaai/Li00K19,kim2019unsupervised}.
Normally, these unsupervised models are trained in a sequential manner.
Our model follows a similar generation paradigm, but has stark differences in model architecture and training approach.
\vspace{-5pt}
\paragraph{Composition Models.} A composition model transforms text encoding into a combinatorial optimization problem, learns and searches for the optimal structure, and encodes the text in a bottom-up manner along a binary tree via a composition function recursively. \newcite{DBLP:journals/nle/MaillardCY19} proposes a CKY-like~\cite{10.5555/1097042,kasami1966efficient,younger1967recognition} encoder, in which high-level constituents are soft-weighted over composed representations of its sub-constituents. \newcite{dblp:conf/naacl/drozdovvyim19} proposes a deep inside-outside encoder~\cite{Baker1979TrainableGF,LARI199035}, enabling the model to learn underlying structures via an auto-encoding objective. 
Recently, a series of studies~\cite{DBLP:conf/acl/HuMWWSZM20,hu-etal-2022-fast,hu2024augmenting} have been conducted to reduce the neural inside encoder complexity from cubic to linear. Our SLM is built on top of state-of-the-art composition modeling techniques, in which we achieve unsupervised learning and enhance training parallelism by taking advantage of the pruned inside-outside encoder~\cite{hu2024augmenting}.

%% file: alg.tex
\section{Methodology}
Given a sentence $\mathbf{x}=[x_1, x_2, \ldots, x_n]$ with $x_i$ 
from a vocabulary $\mathbb{V}$ ($1 \leqslant i \leqslant n$), our goal is to train an SLM without gold trees that can simultaneously generate $\mathbf{x}$ and its syntactic structure. We first introduce the generative architecture of GPST, and then elaborate on how to perform training and inference with the model.

\subsection{Generative Model}\label{sec:model_arc}
GPST generates a sentence and its parse tree from left to right via two types of actions, $\gen$ and $\comp$, along with a stack~\cite{dyer-etal-2015-transition} to maintain partially completed sub-trees during generation. $\gen$ generates a word $x$ and pushes its embedding onto the stack. We denote such an action as $\gen(x)$, with $x \in \mathbb{V}$.
$\comp$ pops the top two elements off the stack, computes their composed representation, and pushes it back to the stack. A major difference in model architecture between GPST and existing unsupervised SLMs, such as URNNG \cite{kim2019unsupervised}, is that GPST makes good use of the architecture of Transformers to parameterize the action probabilities and thus hidden states from previous actions can be directly accessed via self-attention during generation.

\begin{figure}
    \centering
    \includegraphics[width=0.48\textwidth]{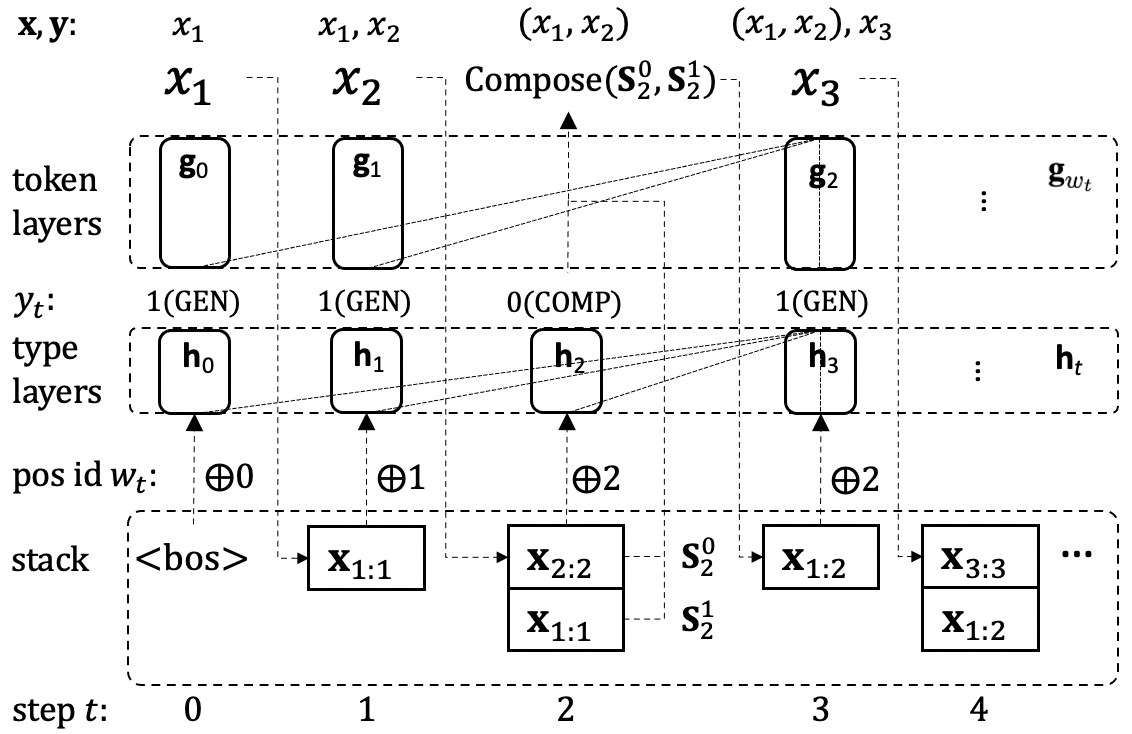}
    \vspace{-5pt}
    \caption{An illustration of the generative process of GPST. $\mathbf{x}_{i:j}$ denotes the sub tree representation spanning from $i$ to $j$. As we use Transformers as the backbone, all previous hidden states are leveraged. At step $t$, the length of historical hidden states is $t$ for the type layers and $w_t$ for the token layers as illustrated with dotted lines for step 3. }
    \label{fig:arch_overview}
    \vspace{-5pt}
\end{figure}

Figure~\ref{fig:arch_overview} illustrates the generative process of GPST. The generative model comprises type layers and token layers, both consisting of multi-layered Transformers. Let us denote the stack at step $t$ as $\mathbf{S}_t$, with $\mathbf{S}_t^0$ and $\mathbf{S}_t^1$ representing the top two elements, respectively. Initially, $\mathbf{S}_0^0$ is set to the embedding of the beginning-of-sentence token (i.e., $\mathbf{\left \langle bos \right \rangle}$ in Figure~\ref{fig:arch_overview}). At each step $t$, $\mathbf{S}_t^0$ along with a position ID $w_t$ is fed into the type layers, yielding a hidden state $\mathbf{h}_t$, which is then utilized to predict the next action type $y_t$
\begin{itemize}[leftmargin=*,noitemsep,nolistsep]
\item If $y_t=0$ ($\comp$), we set $w_{t+1}$ as $w_t$, pop off $\mathbf{S}_t^0$ and $\mathbf{S}_t^1$ from $\mathbf{S}_t$, and compose them using a composition function. 
The composed representation is then pushed back into the stack. In such a case, action $a_t$ at time step $t$ is set to $\comp$. 
\item If $y_t=1$ ($\gen$), we set $w_{t+1}$ as $w_t+1$, feed $\mathbf{h}_t$ to the subsequent token layers, and get an output state $\mathbf{g}_{w_{t}}$ that is used to generate $x_{w_t+1}$. In such a case, we have $a_t=\gen(x_{w_t+1})$.
\end{itemize}
Suppose that $\mathbf{a}_{\mathbf{xy}}$ is the action sequence to generate a sentence $\mathbf{x}$ and its parse tree $\mathbf{y}$, then the joint distribution of $\mathbf{x}$ and $\mathbf{y}$ can be formulated as:
\vspace{-5pt}
\begin{equation}
\small
p(\mathbf{x},\mathbf{y})=p(\mathbf{a}_{\mathbf{xy}})=\prod_{t}p(a_t|a_{<t}),\\
\vspace{-5pt}
\label{eq:auto_regression}
\end{equation}
where $p(a_t|a_{<t})$ is computed by:
\begin{equation*}
\small
\begin{split}
&~~~~~~~~~~~~~~~~~~~~p(\comp|a_{<t}) = p(y_t=0|a_{<t}),\\
&p(\gen(x_{w_t+1})|a_{<t})=p(x_{w_t+1}|y_t=1, a_{<t})p(y_t=1|a_{<t}),\\
&~~~~~~~p(x_{w_t+1}|y_t=1, a_{<t})=\Softmax(\MLP_x(\mathbf{g}_{w_t})),\\
&~~~~~~~~~~~~~~~~~~p(y_t|a_{<t})=\Softmax(\MLP_y(\mathbf{h}_t)).\\
\end{split}
\end{equation*}
$\MLP_y(\cdot)$ and $\MLP_x(\cdot)$ convert inputs to a 2-dim vector and a $\left | \mathbb{V} \right | $-dim vector, respectively. By predicting action types through shallow layers and tokens through deep layers, we can keep the total computational cost close to that of vanilla Transformers.

\subsection{Unsupervised Training}
\label{subsec:unsupervised_training}
\begin{figure*}
    \centering
    \includegraphics[width=1.0\textwidth]{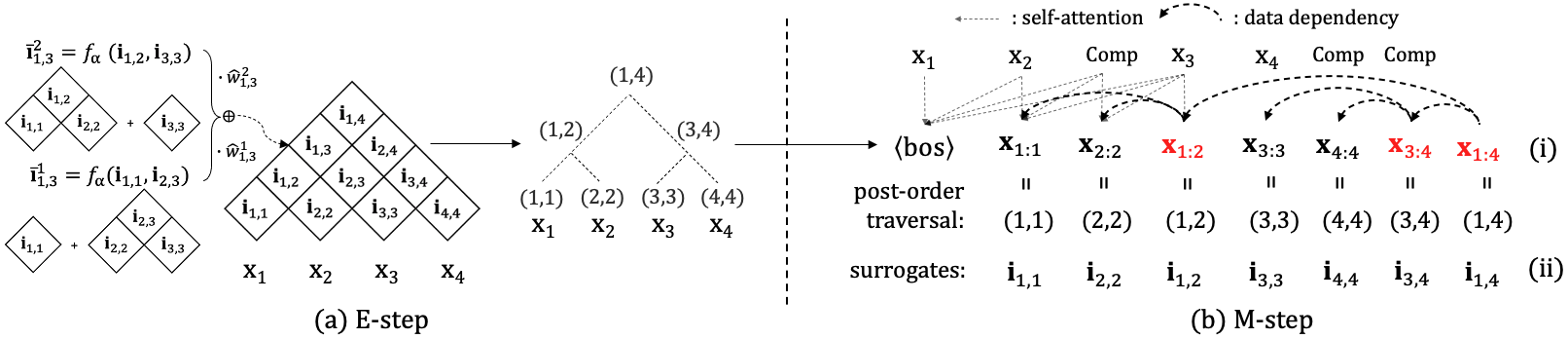}
    \vspace{-5pt}
    \caption{ Illustration of the training process.  
    (a) In the E-step, we induce a parse tree and compute constituent representations.
    (b)(i) Data dependencies within inputs of the generative model.
    (b)(ii) Illustration of representation surrogates. $\mathbf{x}_{i:j}$ denotes the original input representation spanning over $(i,j)$ composed from left to right.
    }
    \vspace{-5pt}
    \label{fig:induction_deduction}
\end{figure*}
How to train an unsupervised SLM effectively and efficiently has always been a challenge.
Existing methods suffer from two issues: asymmetric feedback and inability to train in parallel. The former arises from the uni-directional LM loss, and the latter stems from the inherent data dependency of each composition step on the representations of its sub-constituents from previous steps.
We tackle both issues with an approach similar to hard-EM.
In E-step, we employ a composition model to induce a parse tree through a pruned deep inside-outside encoder.
In M-step, we update both the composition model and the generative model by optimizing a joint objective based on the induced tree. 
The composition and generative model are connected by sharing the same composition function.
Below we present details of the two steps and explain how they tackle the issues mentioned above.

\vspace{-5pt}
\paragraph{E-step.}
During the E-step, the composition model searches for the best parse tree and composes representations through a deep inside-outside auto-encoder 
(DIORA)~\cite{dblp:conf/naacl/drozdovvyim19}. 
In the inside pass, we compute a composed representation $\bar{\mathbf{i}}_{i,j}^k$ and a compatibility score $\bar{a}_{i,j}^k$ for each pair of neighboring constituents $(i,k)$ and $(k+1,j)$.
We then compute each internal span representation $\mathbf{i}_{i,j}$ as a weighted average over all possible pairs of constituents\footnote{A nuance between ours and the original DIORA is that we use the single-step compatibility score $\bar{a}_{i,j}^k$ to estimate the split probability. We empirically find modeling in this way results in better grammar induction performance.}:
\begin{equation*}
\small
\begin{split}
&\bar{\mathbf{i}}_{i,j}^k = f_{\alpha}(\mathbf{i}_{i, k}, \mathbf{i}_{k+1, j})\,,\bar{a}_{i,j}^k = \phi_{\alpha}(\mathbf{i}_{i,k}, \mathbf{i}_{k+1, j})\,,\\
&\,\hat{w}_{i,j}^k = \frac{\exp(\bar{a}_{i,j}^k)}{\sum_{k'=i}^{j-1}\exp(\bar{a}_{i,j}^{k'})}\,,\mathbf{i}_{i,j} = \sum_{k=i}^{j-1}\hat{w}_{i,j}^k\bar{\mathbf{i}}_{i,j}^k\,.\\
\end{split}
\end{equation*}
in which $f_{\alpha}$ and $\phi_{\alpha}$ are formulated in Appendix~\ref{appdx:tree_encoder}.
An illustration to compute $\mathbf{i}_{1,3}$ is given in Figure~\ref{fig:induction_deduction}(a). Analogously, the outside pass computes each outside representation $\mathbf{o}_{i,j}$ in a top-down manner based on bi-directional information outside span $(i,j)$.
To accelerate computation, we use the pruned deep inside-outside encoder~\cite{hu2024augmenting} which achieves linear space complexity and approximately logarithmic parallel time complexity. 
The details of the algorithm and the complete outside pass are presented in Appendix~\ref{appdx:fast_io}.

Note that for a given span $(i,j)$, the best split-point is $k$ with the highest $\bar{a}_{i,j}^k$. Thus, to derive a parse tree,
we can recursively select the best split-points top-down starting from the root span $(1,n)$.

The outside representations of tokens can be used to define an auto-encoding loss (i.e., predicting each token from its outside representation) for the composition model, which is optimized in the M-step:
\vspace{-5pt}
\begin{equation*}
\small
\begin{split}
\mathcal{L}_{ae} = -\frac{1}{n}\sum_{i=1}^{n}\log\frac{\exp(\mathbf{o}^T_{i,i}\mathbf{e}_{\mathbf{x}_i})}{\sum_{k=1}^{\left | \mathbb{V} \right | }{\exp(\mathbf{o}^T_{i,i}\mathbf{e}_{k})}}.
\end{split}
\end{equation*}
where $\mathbf{e}_k$ is the embedding of the $k$-{th} token in the vocabulary. As the auto-encoding loss provides feedback to each token representation from both sides of the token, the asymmetric feedback issue is addressed.

\paragraph{M-step.} 
With the induced tree $\mathbf{y}$, we update the parameters of the composition model and the generative model in a joint manner. Denote the sequence of node spans in post-order as $[(i_0,j_0), (i_1,j_1), ..., (i_{2n-1}, j_{2n-1})]$. The action sequence can be formulated as:
\vspace{-5pt}
\begin{equation*}
\small
\begin{split}
a_t=\left\{\begin{matrix}
\comp~~~~~~\\
\gen(x_{i_t})
\end{matrix}\right.
\,,~\for~
\begin{matrix}
i_t<j_t\\
i_t = j_t
\end{matrix}
.
\end{split}
\end{equation*}

An auto-regression loss can be defined as:
\begin{equation*}
\small
\begin{split}
\mathcal{L}_{ar} = -\log p(\mathbf{x},\mathbf{y}) = -\frac{1}{2n-1}\sum_{t=0}^{2n-1}{\log p(a_t|a_{<t})}.
\end{split}
\end{equation*}

However, even though the action sequence is given, there are still two challenges. First, there are data dependencies within the inputs for the generative model as mentioned earlier and shown in Figure~\ref{fig:induction_deduction}(b), which impedes parallel training. Second, there are no feedforwards from the composition model to the generative model, so the two models are disconnected and hence cannot be trained jointly.
A key insight to tackle these issues is to use the internal span representations $\mathbf{i}_{i,j}$ as surrogates\footnote{It is an approximation because $\mathbf{i}_{i,j}$ is soft-weighted over all $\bar{\mathbf{i}}_{i,j}^k$ with the best split-point at $\hat{k}$. Though $\mathbf{x}_{i,j}$ is composed using the same composition function and with the same split-point $\hat{k}$ during inference, it is in a one-hot manner. However, our experimental results indicate that such an approximation approach has minimal impact on actual inference. } of the inputs $\mathbf{x}_{i:j}$ to the generative model as depicted in Figure~\ref{fig:induction_deduction}(b)(ii).
As the internal span representations are already computed in the E-step, they can be fed into Transformers seamlessly all at once to fully leverage the parallel training ability of the architecture.
Moreover, replacing $\mathbf{x}$ with $\mathbf{i}$ enables the representations computed by the composition model to participate in the generative model, thus the two models are connected and can be jointly optimized via the auto-regression loss. Note that internal span representations do not contain any information outside spans, so there is no information leakage in the uni-directional generative model. 

The final training loss for the M-step combines the auto-regression and auto-encoding losses as:

\begin{equation*}
\small
\mathcal{L}=\mathcal{L}_{ae} + \mathcal{L}_{ar}.
\end{equation*}

We empirically find that the combined loss leads to left-branching bias in parse trees induced by the composition model that is not observed when training with $\mathcal{L}_{ae}$ alone. A possible reason is that left-leaning trees provide more left-side context for each step during generation and thus are reinforced in learning.
To tackle the issue, we stop gradient propagation from $\mathcal{L}_{ar}$ to $a^k_{i,j}$, which means only gradients from $\mathcal{L}_{ae}$ are allowed to be back-propagated along $a_{i,j}^k$. Note that other variables like $\mathbf{i}_{i,j}$ still receive gradients from $\mathcal{L}_{ar}$.

\subsection{Inference}
\label{subsec: inference}
\begin{figure}
    \centering
    \includegraphics[width=0.48\textwidth]{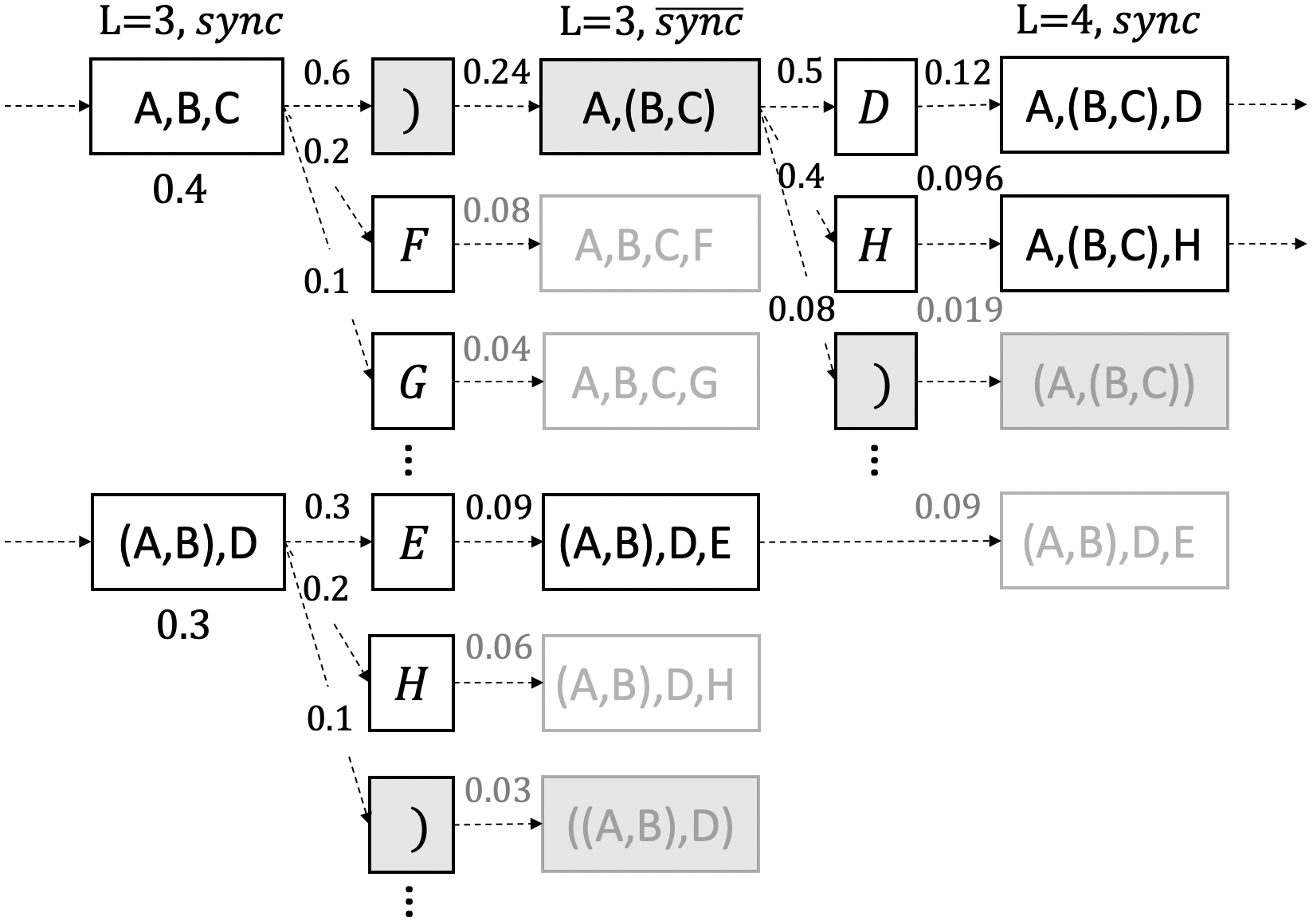}
    \caption{\small An illustration of beam search decoding of size 2. For simplicity, we use ``)" to denote $\comp$ and upper case characters to denote words generated by $\gen(x)$.
    Boxes filled in gray are hypotheses with the last action being $\comp$. Grayed-out boxes are pruned out during beam search.}
    \label{fig:inference}
\end{figure}

The action space of $\gen(x)$ is much larger than that of $\comp$, leading to an imbalance between their probabilities.
~\newcite{stern-etal-2017-effective} point out that during beam search decoding, hypotheses in a beam should be grouped by the length of generated tokens instead of action history, which they refer to as ``word-level search".
However, 
their approach does not guarantee that the top-k next words searched are the optimal ones.
To address the issue, we propose an improved word-level search tailored for our generation paradigm. The core idea is to guarantee that all hypotheses in a beam have the same number of $\gen(x)$ actions. Beams satisfying the condition are marked as $sync$ and otherwise $\overline{sync}$.
Below we depict the entire word-level search process through an example shown in Figure~\ref{fig:inference}:
\vspace{10pt}
\begin{enumerate}[leftmargin=*,noitemsep,nolistsep]
\item Starting with a $sync$ beam, e.g., $A,B,C$ and $(A,B),D$, we estimate the probability distribution of the next action for each hypothesis within it. For each possible action, we compute the probability of the resulting new hypothesis as the product of the probabilities of the current hypothesis and the action. The new hypotheses are pooled and ranked, and the top-k are retained (e.g., $A,(B,C)$ and $(A,B),D,E$).
\item If the current beam contains hypotheses with the last step being $\comp$, e.g., $A,(B,C)$, we continue to explore their next actions, update their probabilities, pool them with other hypotheses in the beam, and rank the top-k, until all the top-k hypotheses have $\gen(x)$ as their last action.
\item End generation upon reaching the length limit or producing an end token; otherwise, go back to step 1.
\end{enumerate}
\vspace{10pt}
This method is also applicable to top-$k$ random sampling~\cite{DBLP:conf/acl/LewisDF18}, or parsing with a given input sentence by simply setting the probabilities of all $\gen(x)$ actions to zeros except for the correct next token.

%% file: experiments.tex
\section{Experiments}
\begin{table*}[htb!]
\small
\begin{center}
\setlength{\tabcolsep}{2pt}
\resizebox{0.98\textwidth}{!}{
\begin{tabular}{lcccccccccc}
Models                             & corpus  & SST2  & COLA  & MRPC(f1)    & QQP(f1)     & QNLI  & RTE   & MNLI-(m/mm) & average & \#param.  \\ \hline \hline
GPT-2$_\text{small}$               & wiki103 & 88.11 & 27.75 & 80.80 & \textbf{85.37} & 83.71 & 53.91 & 75.85/75.77 & 71.41   & 1.0x \\ 
GPST$_\text{small~w/o~grad.stop}$  & wiki103 & 88.11 & 29.09 & 81.16 & 84.98 & 84.62 & 53.19 & 75.87/75.88 & 71.61   &  1.05x   \\ 
GPST$_\text{small~w/o~surrogate}$  & wiki103 & 88.07 & \textbf{29.24} & 80.98 & 85.08 & 84.05 & 52.71 & 76.47/76.36 & 71.62   &  1.05x   \\
GPST$_\text{small}$                & wiki103 & \textbf{88.34} & 28.41 & \textbf{81.21} & 85.33 & \textbf{85.08} & \textbf{56.08} & \textbf{76.60/76.46} & \textbf{72.19}   &  1.05x    \\
\Xhline{1px}
GPT-2$_\text{small}$               & opw & 90.71 & 40.53 & 83.20 & 86.55 & 85.60 & 58.72 & 79.53/79.75 & 75.57   &  1.0x    \\
GPT-2$_\text{small\_13}$           & opw & 91.28 & 44.07 & 83.99 & 86.75 & 85.94 & 58.84 & 79.46/79.82 & 76.27   &  1.1x    \\
GPST$_\text{small}$                & opw & 90.94 & 44.51 & 84.72 & 86.70 & 86.91 & \textbf{64.98} & 79.60/80.15 & 77.31 &  1.05x  \\
GPT-2$_\text{medium}$              & opw & 91.10 & 47.55 & 83.68 & 87.17 & 86.64 & 61.49 & 81.35/81.05 & 77.50   &  2.0x  \\
GPT-2$_\text{medium\_25}$          & opw & 91.55 & 46.81 & 83.43 & 87.27 & 86.99 & 59.92 & 81.19/80.78 & 77.24   &  2.1x    \\
GPST$_\text{medium}$               & opw & \textbf{91.97} & \textbf{50.79} & \textbf{85.69} & \textbf{87.36} & \textbf{87.60} & 64.86 & \textbf{81.80/82.01} & \textbf{79.01} &  2.1x \\
\Xhline{1px} 
\textbf{For Reference}\\ \hline
Ordered-Memory\textsuperscript{\dag}& -- & 90.40 & --    & --/--       & --/--       & --    & --    & 72.53/73.20 & --      &      \\
\hline \hline
\end{tabular}
}
\end{center}
\caption{Evaluation results on GLUE benchmark. We mark out the best result of each group in bold. The results of Ordered-Memory\textsuperscript{\dag}\cite{shen2019ordered} are copied from \newcite{Chowdhury2023beam}. A comparison of parameter sizes can be found in Appendix~\ref{appdx:comp_param_size}.}
\label{tbl:glue}
\end{table*}
To fairly compare GPST and GPT-2, we pre-train both models from scratch on the same corpus with the same setups and comparable parameter sizes. Evaluation is conducted on various language understanding/generation tasks.
Besides, we also evaluate GPST on grammar induction to verify 
to what extent the induced parse trees are consistent with human annotation.

\paragraph{Pre-training Corpus.}
We pre-train models on WikiText-103~\cite{DBLP:conf/iclr/MerityX0S17} and OpenWebText~\cite{Gokaslan2019OpenWeb}, where the two datasets contain $116$ million tokens and $9$ billion tokens, respectively.
The context window size in pre-training is set to $1024$. When a context involves more than one complete sentence, parse trees are induced for each sentence separately.

\paragraph{Hyper-parameters.}
Following GPT-2~\cite{radfordlanguage}, we use $768$/$1024$-dimensional embeddings, a vocabulary size of $30522$, $3072$/$4096$-dimensional hidden layer representations, and $12$/$16$ attention heads for the generative models of GPST$_{\text{small}}$ and GPST$_{\text{medium}}$, respectively. To align with Transformer layer counts in GPT-2, we configure GPST$_{\text{small}}$ with 3 type layers and 9 token layers, and GPST$_{\text{medium}}$ with 3 type layers and 21 token layers, respectively.
We set the input dimension of the composition model to $256$/$512$, and the number of Transformer layers used in the composition function and decomposition function to $4$ and $1$, corresponding to the small and medium setups. To compare with GPT-2 under the same parameter sizes, we provide GPT-2$_{\text{small\_13}}$ and GPT-2$_{\text{medium\_25}}$ as additional baselines, representing models with 13 and 25 layers, respectively.
The token embeddings are down-scaled before being fed into the composition model, and the constituent representations are up-scaled before being fed into GPST. All models are trained on 8 A100 GPUs with a learning rate of 5e-5/1e-4, $8 \times 32 \times 1024$ tokens per step, $5$ billion and $15$ billion total training tokens for WikiText-103 and OpenWebText, respectively. 
\subsection{Understanding Tasks}\label{sec:understanding}
\paragraph{Dataset.}
We evaluate GPST on the GLUE benchmark \cite{wang-etal-2018-glue}, which collects tasks covering a broad range of natural language understanding (NLU) domains.

\paragraph{Evaluation Settings.}
We borrow and minimally modify the fine-tuning paradigm from  \citet{radford2018improving}. Details are described in Appendix~\ref{appdx:finetune}.
As the whole sentence is given, the composition model is utilized to induce the best tree and compose constituent representations as described in the E-step. The constituent representations in the induced tree are gathered in post-order as inputs for the generative model.
We derive two additional baselines GPST$_{\text{w/o~surrogate}}$ and GPST$_{\text{w/o~grad.stop}}$ for ablation study. In GPST$_{\text{w/o~surrogate}}$, all constituent representations of non-terminals are replaced by embeddings of a placeholder \verb|COMP| as in Transformer Grammars~\cite{sartran2022transformer}, and thus there is no interaction between the composition model and the generative model (i.e., they are separately optimized). In GPST$_{\text{w/o~grad.stop}}$, partial gradient stopping is disabled to study the impact of left-leaning trees on downstream tasks.
We run three rounds of fine-tuning with different seeds and report average results (accuracy by default) on the validation sets. 

\paragraph{Results and Discussions.}
Table~\ref{tbl:glue} reports the results on the GLUE benchmark. GPST significantly outperforms GPT-2 in both small and medium setups. 
We find that GPST$_{\text{w/o~grad.stop}}$ and GPST$_{\text{w/o~surrogate}}$ underperform GPST, but are still better than GPT-2. The performance drop of GPST$_{\text{w/o~grad.stop}}$ indicates that poor structures compromise the performance of downstream tasks. GPST$_{\text{w/o~surrogate}}$ is better than GPT-2, implying that as long as induced syntactic structures are utilized, simply replacing non-terminal representations with \verb|COMP| embeddings is also helpful. 
It is, however, worse than GPST, demonstrating that computing representations of non-terminal constituents via an explicit composition function further benefits language understanding.
One more interesting thing we find is that GPST consistently and most significantly outperforms baselines on the RTE task. One possible explanation is that certain relationships in RTE are predicated on negation words such as ``not'', which generally affects high-level semantics through compositions with other phrases. Explicit syntactic composition modeling contributes to a better representation of such cases. 

\begin{table*}[htb!]
\small
\begin{center}
\setlength{\tabcolsep}{2pt}
\resizebox{0.90\textwidth}{!}{
\begin{tabular}{l|c|cccc|cccc|cccc}
\multirow{2}{*}{Models}          & \multirow{2}{*}{\#param.}  &\multicolumn{4}{c|}{XSum} & \multicolumn{4}{c|}{CNN/DailyMail} & \multicolumn{4}{c}{Gigaword} \\ 
                                 &&R-1 & R-2 & R-L & R-AVG & R-1 & R-2 & R-L & R-AVG & R-1 & R-2 & R-L & R-AVG  \\
\hline \hline                          
GPT-2$_{\text{small}}$           &1.0& 29.78 & 9.43 & 23.56 & 20.92 & 35.54 & 14.45 & 24.76 & 24.92 & 32.45 & \textbf{14.84} & 30.37 & 25.88 \\
GPT-2$_{\text{small\_13}}$       &1.1& 29.84 & 9.46 & 23.62 & 20.97 & \textbf{35.78} & 14.58 & 24.92 & \textbf{25.09} & \textbf{32.71} & 14.54 & 30.35 & 25.87 \\
GPST$_{\text{small-w/o~sync}}$   &1.05& 29.44 & 9.09 & 23.20 & 20.58 & 35.63 & 14.57 & 24.93 & 25.04 & 32.34 & 14.69 & 29.98 & 25.67 \\ 
GPST$_{\text{small}}$            &1.05& \textbf{29.86} & \textbf{9.51} & \textbf{23.70} & \textbf{21.02} & 35.52 & \textbf{14.65} & \textbf{25.01} & 25.06 & 32.53 & 14.76 & 30.37 & \textbf{25.89} \\ 
\hline 
GPT-2$_{\text{medium}}$          &2.0& 31.91 & 11.11 & 25.28 & 22.76 & 37.18 & 15.23 & 25.59 & 26.00 & 33.13 & 15.27 & 30.85 & 26.42 \\ 
GPT-2$_{\text{medium\_25}}$      &2.1& 31.95 & 11.17 & 25.35 & 22.82 & 37.13 & 15.26 & 25.59 & 25.99 & \textbf{33.49} & 15.27 & \textbf{31.28} & \textbf{26.68} \\ 
GPST$_{\text{medium~w/o~sync}}$  &2.1& 31.66 & 10.91 & 25.16 & 22.58 & 37.07 & 15.45 & 25.69 & 26.07 & 32.83 & 15.06 & 30.59 & 26.16 \\ 
GPST$_{\text{medium}}$           &2.1& \textbf{31.96} & \textbf{11.31} & \textbf{25.58} & \textbf{22.95} & 37.18 & \textbf{15.69} & \textbf{26.00} & \textbf{26.29} & 33.19 & 15.27 & 30.91 & 26.46 \\ 
\hline \hline
\end{tabular}
}
\end{center}
\vspace{-10pt}
\caption{Abstractive summarization results.}
\vspace{-5pt}
\label{tbl:summary}
\end{table*}
\begin{table}[htb!]
\small
\begin{center}
\setlength{\tabcolsep}{2pt}
\resizebox{0.5\textwidth}{!}{
\begin{tabular}{lccccccc}
Models                          & Agr.  & C.E.  & G.P.E. & C.S.E. & Lcs.  & L.D.D. & avg   \\ \hline \hline 
\textbf{WikiText-103}\\ \hline    
GPT2$_{\text{small}}$           & 50.88 & 73.21 & 77.88  & 97.83  & 33.95 & 65.98  & 66.62 \\  
GPST$_{\text{small}}$           & 59.65 & 73.21 & 87.10  & 97.83  & 57.89 & 64.78  & 73.41 \\ 
\Xhline{1px}
\textbf{OpenWebText}\\ \hline                                     
GPT$_{\text{small}}$            & 78.95 & 87.50 & 85.22  & 97.83  & 71.58 & 78.65  & 83.29 \\ 
GPST$_{\text{small}}$           & 77.19 & 85.71 & 94.54  & 96.74  & 68.95 & 72.38  & 82.59 \\ 
GPT$_{\text{medium}}$           & 64.91 & \textbf{94.64} & 86.41  & \textbf{98.91}  & 73.42 & \textbf{79.38}  & 82.95 \\ 
GPST$_{\text{medium}}$          & \textbf{85.96} & 85.71 & \textbf{95.04}  & 94.57  & \textbf{83.68} & 78.17  & \textbf{87.19} \\ 
\Xhline{1px}
\multicolumn{4}{l}{\textbf{For Reference~(Models with gold trees)}} \\ \hline
TG                              & 69.7  &  88.4 &  90.4  & 95.6   & 78.1  &  77.9  &  83.35 \\
Pushdown Layers                 & 79.0  &  92.0 &  84.2  & 100.0  & 77.8  &  77.5  &  85.08 \\
\hline \hline
\end{tabular}
}
\end{center}
\vspace{-10pt}
\caption{ Syntactic generalization results. For reference, we list the results of models with gold trees from \citet{sartran2022transformer} and \citet{murty2023pushdown}.}
\vspace{-15pt}
\label{tbl:sg}
\end{table}

\subsection{Generation Tasks}

\subsubsection{Abstractive Summarization}
\paragraph{Datasets.}
We conduct experiments on three summarization datasets: BBC extreme (XSum) \cite{narayan-etal-2018-dont}, CNN and DailyMail \cite{DBLP:conf/conll/NallapatiZSGX16}, and Gigaword \cite{napoles-etal-2012-annotated} to assess the performance of GPST in terms of language generation abilities. Statistics of the datasets are presented in Table \ref{tbl:summarydataset} in Appendix~\ref{appdx:dataset}.

\vspace{-5pt}
\paragraph{Evaluation Settings.} 
For XSum and CNN/DailyMail, we truncate the documents and their summaries to 900 and 100 tokens respectively, and concatenate them with short prompt \verb|Summary:|. For Gigaword, the truncating thresholds of documents and summaries are set to 400 and 120 respectively, following the settings of \citet{rothe-etal-2020-leveraging}.
Considering the complexity of the generation task, we primarily evaluate the models pre-trained on OpenWebText.
More details are described in Appendix~\ref{appdx:sum_finetune}. We apply the word-level search described in \S\ref{subsec: inference} to top-k random sampling for GPST, except for models with $\text{w/o~sync}$ which only uses naive action-level beam search. ROUGE \cite{lin2003automatic} is employed as the evaluation metric.

\subsubsection{Syntactic Generalization}
\paragraph{Datasets.}
The syntactic generalization task \cite{hu-etal-2020-systematic} collects 34 test suites to assess syntactic generalizability of the models. The test suites are grouped into 6 circuits: Agreement (Agr.), Center Embedding (C.E), Garden-Path Effects (G.P.E), Cross Syntactic Expectation (C.S.E.), Licensing (Lcs.) and Long-Distance Dependencies (L.D.D.).
\vspace{-5pt}
\paragraph{Evaluation Settings.} 
We evaluate models on syntactic generalization test suites by comparing surprisals \cite{hale2001probabilistic} without fine-tuning, as required by \citet{hu-etal-2020-systematic}.
Surprisal: $S(w|C)=-\log_2p(w|C)$ is defined as negative log conditional probabilities of a sub-sentence $w$ given the left-side context $C$. 
In detail, when we apply word-level search with beam size $b$ to do left-to-right parsing with a given input, we temporarily store $b$ best hypotheses with their probability $p(\mathbf{x}_{<t}, \mathbf{y}_{<n(t)})$ at each token position $t$, in which $\mathbf{y}_{<n(t)}$ refers to the current latent structure before generating $x_{t}$. We marginalize $\mathbf{y}_{<n(t)}$ out of $p(\mathbf{x}_{<t}, \mathbf{y}_{<n(t)})$ by summing up all the probabilities of the $b$ best hypotheses. Finally, we obtain the surprisal of a sub-sentence with starting position $s$ and ending position $e$ as $S(w|C)=-\log{p(x_{<e})} + \log{p(x_{<s-1})}.$

To align with \citet{murty2023pushdown} and \citet{sartran2022transformer}, we set beam size $b$ to 300.
\subsubsection{Results and Discussions}
Table~\ref{tbl:summary} and \ref{tbl:sg} report the results of summarization and syntactic generalization tasks. Overall, the performance of GPST is comparable to GPT, with a slight advantage. 
One possible reason why the advantage of GPST on generalization tasks is not as significant as that on GLUE is the discrepancies between training and inference. During training, the constituent representations are computed via the inside algorithm, where the representations are soft-weighed over composed representations of valid sub-constituents. However, during inference, constituent representations are composed of the top two elements in the stack, which is a one-hot version of the inside algorithm. This issue could potentially be resolved using a hard inside-outside algorithm~\cite{DBLP:conf/emnlp/DrozdovRCOIM20}, which we may explore in our future work. Despite the discrepancies, our performance still slightly surpasses that of GPT-2, which adequately demonstrates the potential of GPST in generation tasks.
One more interesting thing is that GPST$_{\text{medium}}$ even outperforms baselines with gold trees in the syntactic generalization task, and the results of all GPSTs maintain a lead on Garden-Path Effect. Note that the results have a large variance due to the relatively small size of the evaluation set, e.g., GPT$_{\text{medium}}$ even under-performs GPT$_\text{small}$.
However, the results still imply that unsupervised syntactic LMs have reached a critical point where they can surpass approaches reliant on gold trees.

\subsection{Grammar Induction}
\paragraph{Baselines \& Dataset.}
We select baselines that report unsupervised left-to-right parsing results: Neural variational (NV) approaches~\cite{DBLP:conf/aaai/Li00K19} and PRPN~\cite{DBLP:conf/iclr/ShenLHC18}. For reference, we also select some baselines performing parsing requiring whole sentence visible: URNNG \cite{kim2019unsupervised}, C-PCFG \cite{kim-etal-2019-compound}, DIORA \cite{dblp:conf/naacl/drozdovvyim19}, ON-LSTM~\cite{DBLP:conf/iclr/ShenTSC19}, Fast-R2D2 \cite{hu-etal-2022-fast} and GPST using the deep inside algorithm. 
We report their performance on PTB \cite{marcus-etal-1993-building}.
Besides, we also report results of GPST$_{\text{w/o~grad.stop}}$ and GPST$_{\text{w/o~surrogate}}$ for checking the gains from partial gradient stopping and joint pre-training achieved by the representation surrogate.

\paragraph{Evaluation Settings.}
We continue to fine-tune all models on the training set of PTB for 10 epochs with batch size set to 32 after pre-training. Since GPST takes word pieces as inputs, we provide our model with word-piece boundaries as non-splittable spans to align with models with word-level inputs. 
We apply different inference algorithms to grammar induction. For the inside algorithm, we directly evaluate the parse tree induced by the composition model. For the left-to-right parsing, we apply improved word-level search described in \S\ref{subsec: inference} with a beam size of 20 to parse the given text, except for GPST$_\text{w/o~sync}$ which employs action-level $\text{sync}$ beam search for parsing.
We adopt sentence-level unlabeled $F_1$ as the evaluation metric, with the same setup as ~\citet{kim-etal-2019-compound} where punctuations are discarded and words are lowercased. We evaluate the checkpoints from all epochs on the validation set, pick the best one, and then report its performance on the test set. 

\begin{table}[htb!]
\centering
\vspace{-5pt}
\resizebox{0.45\textwidth}{!}{
\begin{tabular}{lccc}
Models                       & corpus  &left-to-right & F1    \\ \hline \hline
NV(unsupervised)             & WSJ     &yes  & 29.0  \\
NV($+$linguistic rules)      & WSJ     &yes  & 42.0  \\
PRPN                         & WSJ     &yes  & 37.4  \\
GPST$_\text{small~w/o~sync}$ & wiki103 &yes  & 43.64 \\ 
GPST$_\text{small}$          & wiki103 &yes  & \textbf{55.25} \\
GPST$_\text{small~w/o~sync}$  & opw     &yes   & 43.09 \\ 
GPST$_\text{small}$           & opw     &yes   & 51.40 \\ 
GPST$_\text{medium~w/o~sync}$ & opw     &yes   & 43.37 \\ 
GPST$_\text{medium}$          & opw     &yes   & 54.71 \\ 
\Xhline{1px}
\textbf{For Reference}\\ \hline
GPST$_\text{small~w/o~grad.stop}$ & wiki103 & no & 42.46 \\
GPST$_{\text{small~w/o~surrogate}}$ & wiki103 & no & 50.27 \\
GPST$_\text{small}$               & wiki103 & no & 57.46 \\ 
GPST$_\text{small}$               & opw     & no & 53.95 \\ 
GPST$_\text{medium}$              & opw     & no & 56.27 \\ 
\hline 
URNNG                             & WSJ     & no & 45.4  \\
ON-LSTM                           & WSJ     & no & 47.4  \\
C-PCFG                            & WSJ     & no & 55.2  \\
DIORA                             & WSJ     & no & 55.7  \\
Fast-R2D2                         & wiki103 & no & 57.2  \\ 
Oracle                            &   ---   & ---   & 84.3  \\
\hline \hline
\end{tabular}
}
\vspace{-5pt}
\caption{Results on unsupervised left-to-right parsing.}
\vspace{-15pt}
\label{tbl:grammar_induction}
\end{table}
\paragraph{Results and Discussions.}
There are several observations from the results shown in Table~\ref{tbl:grammar_induction}. First and foremost, we find that our unsupervised left-to-right parsing achieves comparable performance with the bi-directional inside algorithm, significantly surpassing previous left-to-right grammar induction baselines. Such results indicate the structures generated by GPST are meaningful and consistent with those from humans. Secondly, a larger pre-training corpus may not necessarily bring improvement. A plausible explanation is that OpenWebText, mixed with more non-natural text such as URLs, introduces additional noise, leading to a performance drop. The results indicate the importance of high-quality corpora for structural learning. Thirdly, the performance of GPST$_\text{w/o~surrogate}$ infering with the inside algorithm drops a lot. We suppose the main reason is that disabling the representation surrogate prevents the composition model from receiving long-term feedback from tokens on the right introduced by the auto-regression loss.
Lastly, the performance decline of GPST$_\text{w/o~grad.stop}$ corroborates the impact of asymmetric loss on structural learning. 
We attach trees parsed by GPST in Appendix~\ref{appdx:case_study} for case studies.

\subsection{Training Efficiency}
Finally, we conduct a fair comparison of training efficiency with other unsupervised SLMs. We keep the model sizes and memory usage comparable and the training tokens the same. We report their time consumption in Table~\ref{tbl:speed_comparison}, from which we can observe the huge advantage of GPST over the baselines in terms of efficiency. 
\begin{table}[htb!]
\small
\vspace{-10pt}
\begin{tabular}{ll|cccc}
      &          & \multicolumn{4}{c}{sentence length} \\
      & \#param. & 128    & 256    & 512    & 1024   \\ \hline \hline
GPST  & 24M      & 1x     & 1x     & 1x     & 1x     \\
URNNG & 23M      & 130.6x & 955.3x & n/a    & n/a    \\
OM    & 28M      & 2.0x   & 9.2x   & 25.4x  & 63.3x  \\ \hline \hline
\end{tabular}
\vspace{-5pt}
\caption{Training acceleration on the same number of tokens.}
\label{tbl:speed_comparison}
\vspace{-10pt}
\end{table}

%% file: conclusion.tex
\section{Conclusion}
In this paper, we propose an unsupervised approach to train GPST at scale efficiently. A key insight of our work is to guide the left-to-right structural learning with symmetric supervision such as an auto-encoding loss, which can receive feedback from both sides. A key technical contribution is that we propose the representation surrogate which enables joint training of all components in parallel. 
Besides, the composition model of GPST can be regarded as an enhancement to the conventional embedding layer, which provides context-invariant embeddings of various granularities beyond token embeddings.
Our experiment results show the superiority of GPST on language understanding, generation, and left-to-right grammar induction, which demonstrate the potential of GPST as a foundational architecture for large language models.

%% file: limitation.tex
\section{Limitation}
Despite GPST achieving a multiple-fold acceleration compared to previous syntactic language models, it still requires 1.5 to 5 times the training time compared to vanilla GPTs.
The more layers there are in the type/token layers, the lower the overall time multiplier becomes.
The additional training time comes from the composition model which only accounts for one-tenth of the overall model parameters. Due to the unpredictable memory overhead at each step, many fragmented memories are generated, resulting in PyTorch having to spend extra time cleaning up the memory cache periodically. Meanwhile, our implementation is quite naive, without any operator fusion or hardware-aware implementation. Thus there should be multiple potential ways to further reduce the time consumption of the composition model in the future.

\section{Acknowledgement}
This work was supported by Ant Group through the CCF-Ant Research Fund.

%% file: appendix.tex
\begin{figure*}[htb!]
  \centering
  \includegraphics[width=.8\textwidth, clip]{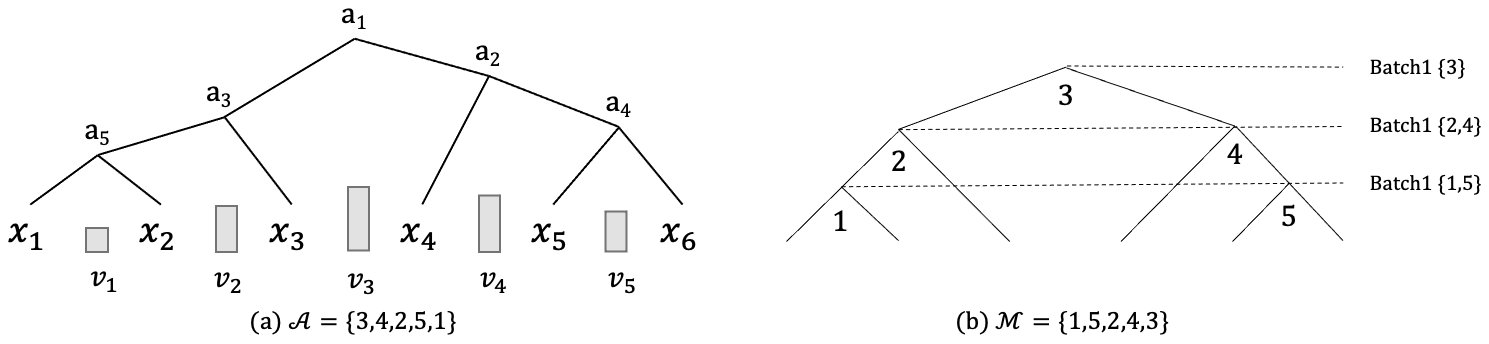}
  \caption{Fast encoding follows the order given by a top-down parser, with the merging order $\mathcal{M}$ being the reverse order of the split point sequence $\mathcal{A}$. $x_i$ denotes the $i_{th}$ token in a sentence of length 6. Numbers in $\mathcal{A}$ and $\mathcal{M}$ denote the indices of the split/merge point between tokens. $v_j$ denotes the split score of $j_{th}$ split point, predicted by the top-down parser. }
  \label{fig:merge_order}
\end{figure*}
\begin{figure*}[htb!]
  \centering
  \includegraphics[width=.98\textwidth, clip]{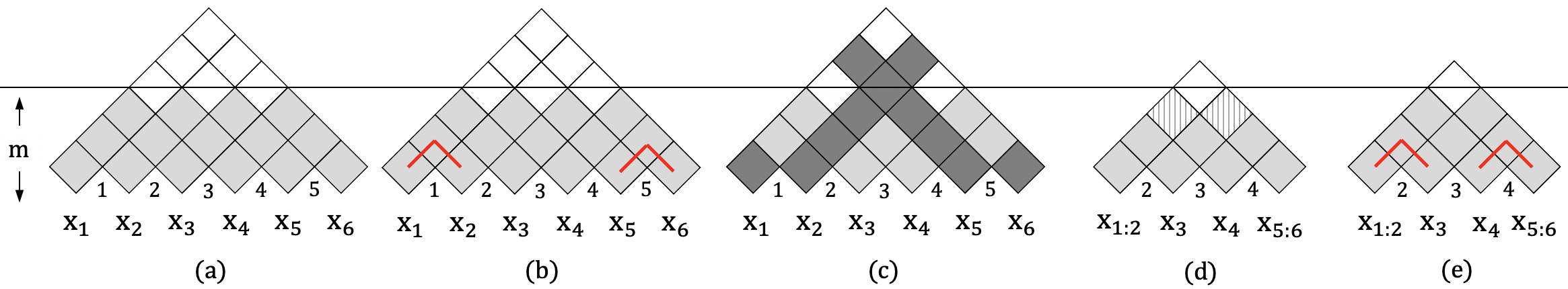}
  \caption{The initial step of encoding in $O(\log n)$ steps. The numbers in blue correspond to the indices of the split points introduced in Figure \ref{fig:merge_order}.}
  \label{fig:logn_io}
\end{figure*}
\subsection{Pruned inside-outside algorithm}\label{appdx:fast_io}
Fast-R2D2 \cite{hu-etal-2022-fast} introduces a pruned variant of the inside algorithm that reduces its complexity from $O(n^3)$ to $O(n)$ in both space and time. Building on this, ReCAT \cite{hu2024augmenting} extends the pruning method to the inside-outside algorithm, and further enables it to complete in approximate $\log n$ steps,
whose key idea is to prune out unnecessary cells in the chart-table and encode cells in different rows simultaneously.
The main idea of the pruning process is to decide which two spans should be merged at each step during the inside pass and prune out cells that would break the non-splittable span. 
An unsupervised top-down parser is applied to determine the merge order of spans. Given a sentence $\mathbf{x}=[x_1, x_2, \ldots, x_n]$, the top-down parser assigns each split point a score $v_i$ s.t. $1\leq i \leq n-1$, and recursively split the sentence into two in the descending order of the scores shown in Figure \ref{fig:merge_order} (a). 
Hence, the reverse order of the split points could be used to decide which cells to merge. 
Specifically, the pruned inside-outside algorithm works as follows:
\setlist[enumerate,1]{start=0}
\begin{enumerate}[leftmargin=*,noitemsep,nolistsep]
    \item Prepare merge batches according to the height of merge points in the induced tree, with the lowest merge points in the first batch, as illustrated in Figure \ref{fig:merge_order}(b).
    \item Merge each pair of adjacent cells into one according to the current merge batch. For example, in Figure \ref{fig:logn_io}(b), at merge point 1, we merge $x_1$ and $x_2$ into $x_{1:2}$; at merge point 5, we merge $x_5$ and $x_6$ into $x_{5:6}$.
    \item Remove all conflicting cells that would break the now non-splittable span from Step 1, e.g., the dark cells in Figure \ref{fig:logn_io}(c), and reorganize the chart table much like in the Tetris game as in (d).
    \item Encode the cells that just descend to height $m$ and record their valid splits in $\mathcal{K}$, e.g., the cells highlighted with stripes in Figure \ref{fig:logn_io}(d) with valid splits $\{2, 3\}$ for span $(1,4)$ and $\{3,4\}$ span $(3, 6)$. Go back to Step 1 until no blank cells are left.
\end{enumerate}
Therefore, the entire inside process can be completed within steps equal to the height of the tree.
Using the valid splits $\mathcal{K}$ recorded for each cell during the pruning process, we now have the new inside state transition equation as:

{
\vspace{-10pt}
\small
\begin{gather*}
\bar{a}_{i,j}^k = \phi_{\alpha}(\mathbf{i}_{i,k}, \mathbf{i}_{k+1, j})\,,
\bar{\mathbf{i}}_{i,j}^k = f_{\alpha}(\mathbf{i}_{i, k},\mathbf{i}_{k+1, j})\,,\\
\hat{w}_{i,j}^k = \frac{\exp(\bar{a}_{i,j}^k)}{\sum_{k'\in\mathcal{K}_{i,j}}\exp(\bar{a}_{i,j}^{k'})}\,,
\mathbf{i}_{i,j} = \sum_{k\in\mathcal{K}_{i,j}}\hat{w}_{i,j}^k\bar{\mathbf{i}}_{i,j}^k\,.
\end{gather*}
}
where $\mathcal{K}_{i,j}$ is the valid splits set for span $(i,j)$.
According to $\mathcal{K}$, we can obtain a mapping from a span to its immediate sub-spans. By reversing such mapping, we get a mapping from a span to its valid immediate parent spans denoted as $\mathcal{P}$, which records the non-overlapping endpoint $k$ in the parent span $(i,k)$ or $(k,j)$ for a given span $(i,j)$.

Thus, for the outside pass, we have:

\begin{equation*}
\small
\begin{split}
&\bar{\mathbf{o}}_{i,j}^k=\left\{\begin{matrix} 
f_{\beta}(\mathbf{o}_{i,k}, \mathbf{i}_{j+1,k}) & \text{ if }k>j \\
f_{\beta}(\mathbf{o}_{k,j}, \mathbf{i}_{k,i-1}) & \text{ if }k<i
\end{matrix}\right. \,,\,\\
&\bar{b}_{i,j}^k=\left\{\begin{matrix}
\phi_{\beta}(\mathbf{o}_{i,k}, \mathbf{i}_{j+1,k})& \text{ if }k>j \\
\phi_{\beta}(\mathbf{o}_{k,j}, \mathbf{i}_{k,i-1})& \text{ if }k<i
\end{matrix}\right. \,,\\
&\check{w}_{i,j}^k=\frac{\exp(\bar{b}_{i,j}^k)}{\sum_{k' \in \mathcal{P}_{i,j}}^{}\exp(\bar{b}_{i,j}^{k'})}\,,\mathbf{o}_{i,j}=\sum_{k \in \mathcal{P}_{i,j}}^{}\check{w}_{i,j}^k\bar{\mathbf{o}}_{i,j}^k\,.
\end{split}
\vspace{-5pt}
\end{equation*}

We optimize the top-down parser jointly at the M-step with GPST. Given the parse tree $\mathbf{y}$ induced at the E-step, we maximize $p(\mathbf{y}|\mathbf{x};\Theta)$ for the top-down parser, 
whose parameters are denoted as $\Theta$.
As shown in Figure~\ref{fig:merge_order}(a), at $t$ step, the span corresponding to a given split $a_t$ is determined, which is denoted as $(i^t, j^t)$.
Thus we can minimize the negative log-likelihood of the parser as follows:

{
\vspace{-10pt}
\small
\begin{gather*}
p(a_{t}|\mathbf{x},\Theta) = \frac{\exp(v_{a_{t}})}{\sum_{k=i^t}^{j^t-1}\exp(v_{k})} \,, \\
\mathcal{L}_p=-\log p(\mathbf{y}|\mathbf{x};\Theta) = -\sum_{t=1}^{n-1} \log p(a_{t}|\mathbf{x};\Theta).
\label{eq:parser_log_likelihood}
\end{gather*}
}

We notice that the steps to finish the pruned inside-outside algorithm depend on the highest tree in a batch, thus any extremely skewed tree may result in a significant increase in time consumption. A straightforward approach to reduce the maximum height of parse trees in a batch is to introduce a height penalty. During the inside pass, the weighted tree height of span $(i,j)$ could be computed as:

{
\vspace{-10pt}
\small
\begin{gather*}
\bar{h}_{i,j}^k = \max(h_{i,k}, h_{k+1,j}) + 1, h_{i,j} = \sum_{k\in\mathcal{K}_{i,j}}^{}{\hat{w}_{i,j}^{k}\bar{h}_{i,j}^{k}}
\end{gather*}
}
To minimize the impact of height penalties on grammar induction, we set a threshold $H_{thrs}$ which is 15 by default. Only trees that exceed this threshold will be affected.

{
\vspace{-10pt}
\small
\begin{gather*}
\mathcal{L}_{h}=\frac{1}{n}\max(h_{1,n}-H_{thrs}, 0)
\end{gather*}
}
Thus the final auto-encoding objective is:

{
\vspace{-10pt}
\small
\begin{gather*}
\mathcal{L}_{ae}^* = \mathcal{L}_{ae} + \mathcal{L}_h + \mathcal{L}_{p}
\end{gather*}
}

\subsection{Composition function and score function}\label{appdx:tree_encoder}
\begin{figure}[htb!]
    \centering
    \includegraphics[width=0.48\textwidth]{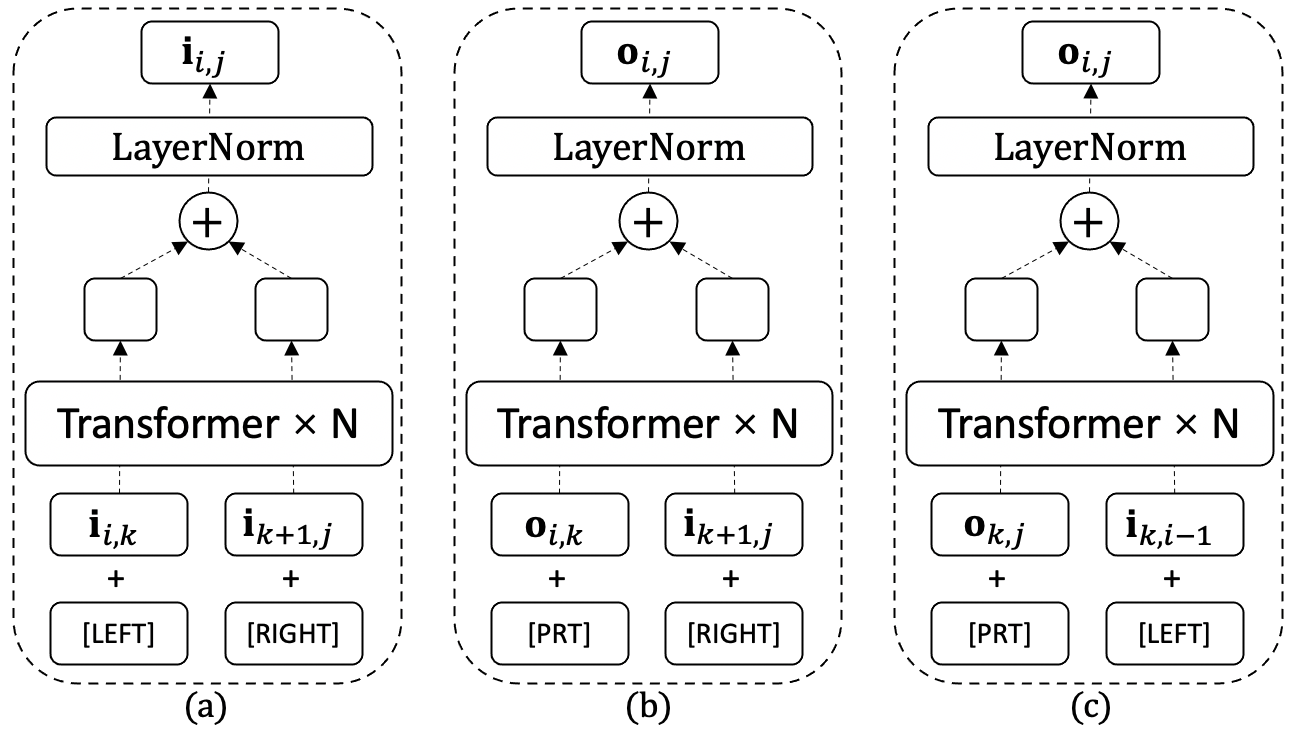}
    \caption{\small Model illustrations for the composition and decomposition functions.}
    \label{fig:comp_func}
\end{figure}
We borrow the idea from ~\newcite{DBLP:conf/acl/HuMWWSZM20} to use Transformers as the backbone of the composition function $f_\alpha$. As shown in Figure~\ref{fig:comp_func}(a), composition function $f_\alpha$ takes left/right constituent representations $\mathbf{i}_{i,k}$/$\mathbf{i}_{k+1,j}$ along with their role embeddings \verb|[LEFT]|/\verb|[RIGHT]| into N-layered Transformers as inputs, passes the summation of their corresponding outputs through a layer normalization layer to get the composed representation. Decomposition function $f_\beta$ works analogously as shown in Figure~\ref{fig:comp_func}(b) and (c), with \verb|[PRT]| as the role embedding for parents.

We define the score function $\phi_\alpha$ as:
\begin{equation}
\small
\phi_\alpha(\mathbf{l},\mathbf{r})=\MLP^l_\alpha(\mathbf{l})^T \MLP^r_\alpha(\mathbf{r})/\sqrt{d}
\end{equation}
where $\mathbf{l}$ and $\mathbf{r}$ are representations  for left/right constituents. $\MLP^l_\alpha$ and $\MLP^r_\alpha$ are used to capture syntactic features from the left and right inputs, which convert inputs to $d$-dimensional vectors.
Analogously, $\phi_\beta$ is defined as:

{
\small
\vspace{-10pt}
\begin{align*}
&\phi_\beta(\mathbf{p},\mathbf{l})=\MLP^p_\beta(\mathbf{p})^T \MLP^l_\beta(\mathbf{l})/\sqrt{d}\\
&\phi_\beta(\mathbf{p},\mathbf{r})=\MLP^p_\beta(\mathbf{p})^T \MLP^r_\beta(\mathbf{r})/\sqrt{d}
\end{align*}
}
where $\mathbf{p}$ is the outside representation of a parent. $\MLP^l_\beta$, $\MLP^r_\beta$, and $\MLP^p_\beta$ are used to capture features from left/right children and parents respectively.

\subsection{Glue fine-tuning}\label{appdx:finetune}
In detail, we append a \verb|CLS| token after the input sequence and then feed the hidden states of the \verb|CLS| tokens to a linear layer as the logits for classification. An additional cross-entropy loss along with the pre-training objective is used during fine-tuning. 

\subsection{Summarization dataset statistics} 
\label{appdx:dataset}
BBC extreme (XSum) comprises 204k document-summary pairs for single-sentence summarization of long documents. CNN and DailyMail (CNN/DailyMail) contains 287k training pairs, each consisting of a document annotated with highlights. Gigaword focuses on sentence summarization with 3.8M sentence-summary training pairs conversely. We organize the statistics in Table~\ref{tbl:summarydataset}.
\begin{table}[htb!]
\begin{center}
\setlength{\tabcolsep}{2pt}
\begin{tabular}{l|c|c|c}
             & XSum  & CNN/DailyMail & Gigaword \\ \hline \hline
Training Set & 204k  & 287k          & 3.8M     \\
Test Set     & 11.3k & 11.5k         & 1.95k    \\ \hline \hline
\end{tabular}
\end{center}
\caption{Detailed statistics for summarization datasets.}
\label{tbl:summarydataset}
\end{table}

\subsection{Summarization fine-tuning}\label{appdx:sum_finetune}
We fine-tune for 15 epochs with a batch size of 16 on XSum and CNN/DailyMail datasets. For Gigaword, we fine-tune for 10 epochs with a batch size of 64.
Top-k random sampling with $k=2$ is used as the basic inference method as suggested in GPT-2~\cite{radfordlanguage}.

\subsection{Comparison of parameter sizes}
\label{appdx:comp_param_size}

\begin{table}[htb!]
\begin{center}
\resizebox{0.5\textwidth}{!}{
\setlength{\tabcolsep}{2pt}
\begin{tabular}{l|c|c|c|c}
model & MLP  & QKV linear & score linear & total \\ \hline \hline
comp. func. & 256*1024*2 & 256*256*3 & 256*128*2  & 786,432     \\
GPT-2    & 768*3072*2 & 768*768*3 & --- & 6,488,064 \\ \hline \hline
\end{tabular}
}
\end{center}
\caption{The comparison of parameter size of a single layer in the composition function of GPST$_\text{small}$ and GPT-2$_\text{small}$.}
\label{tbl:parameter_details}
\end{table}

\subsection{Author Contributions Statement}\label{appdx:contribution_statement}
We list our contributions below:
\begin{itemize}[leftmargin=*,noitemsep,nolistsep]
\item Xiang Hu: proposing the unsupervised training approach, implementing GPST, pre-training and decoding algorithms, and paper writing.
\item Pengyu Ji: GPST fine-tuning, experiments code, running experiments, and paper writing.
\end{itemize}

\subsection{Case studies} \label{appdx:case_study}
Please refer to the following pages.

\begin{figure*}[!htb]
    \centering
    \includegraphics[width=\textwidth]{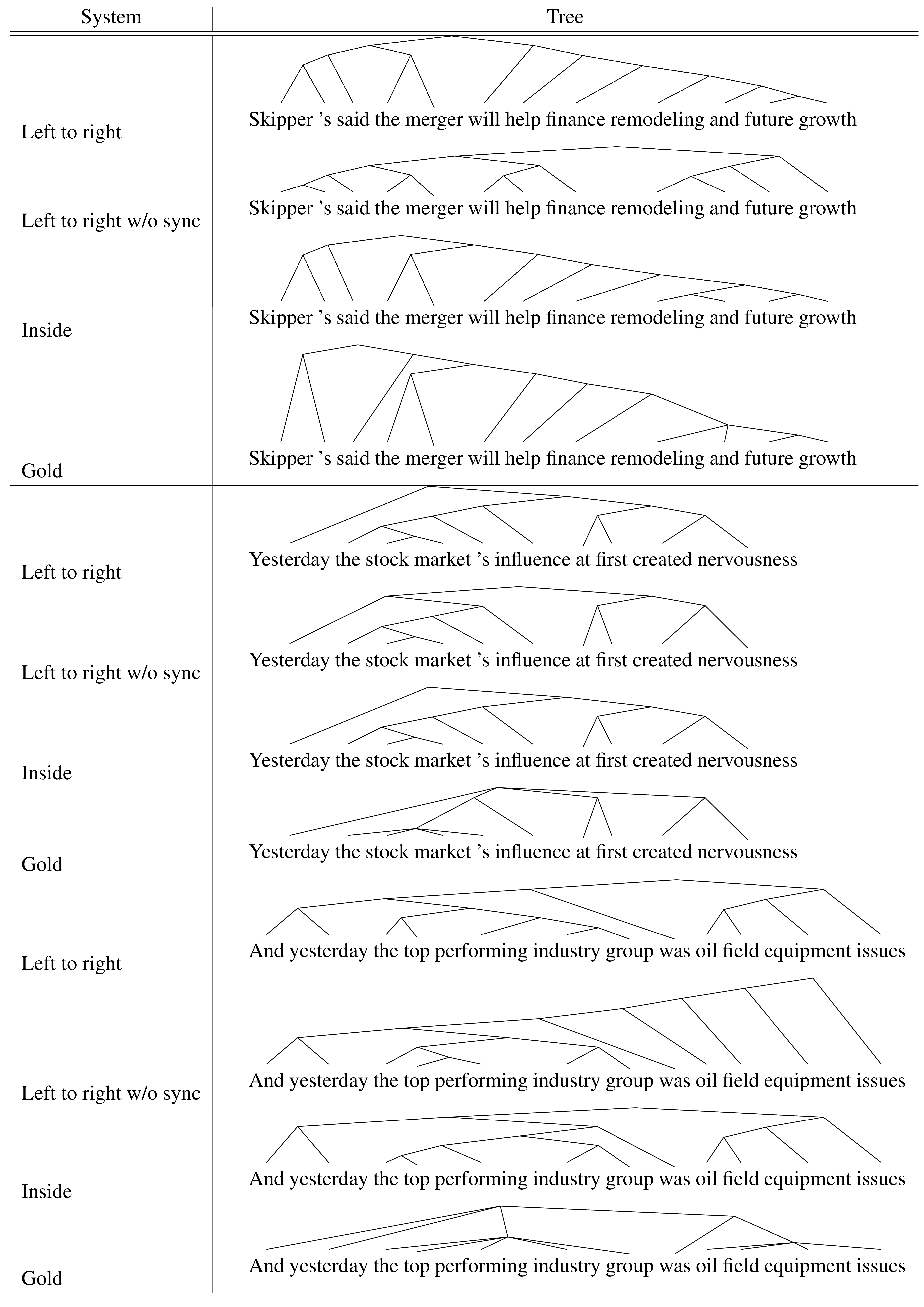}
\end{figure*}

\begin{figure*}[!htb]
    \centering
    \includegraphics[width=\textwidth]{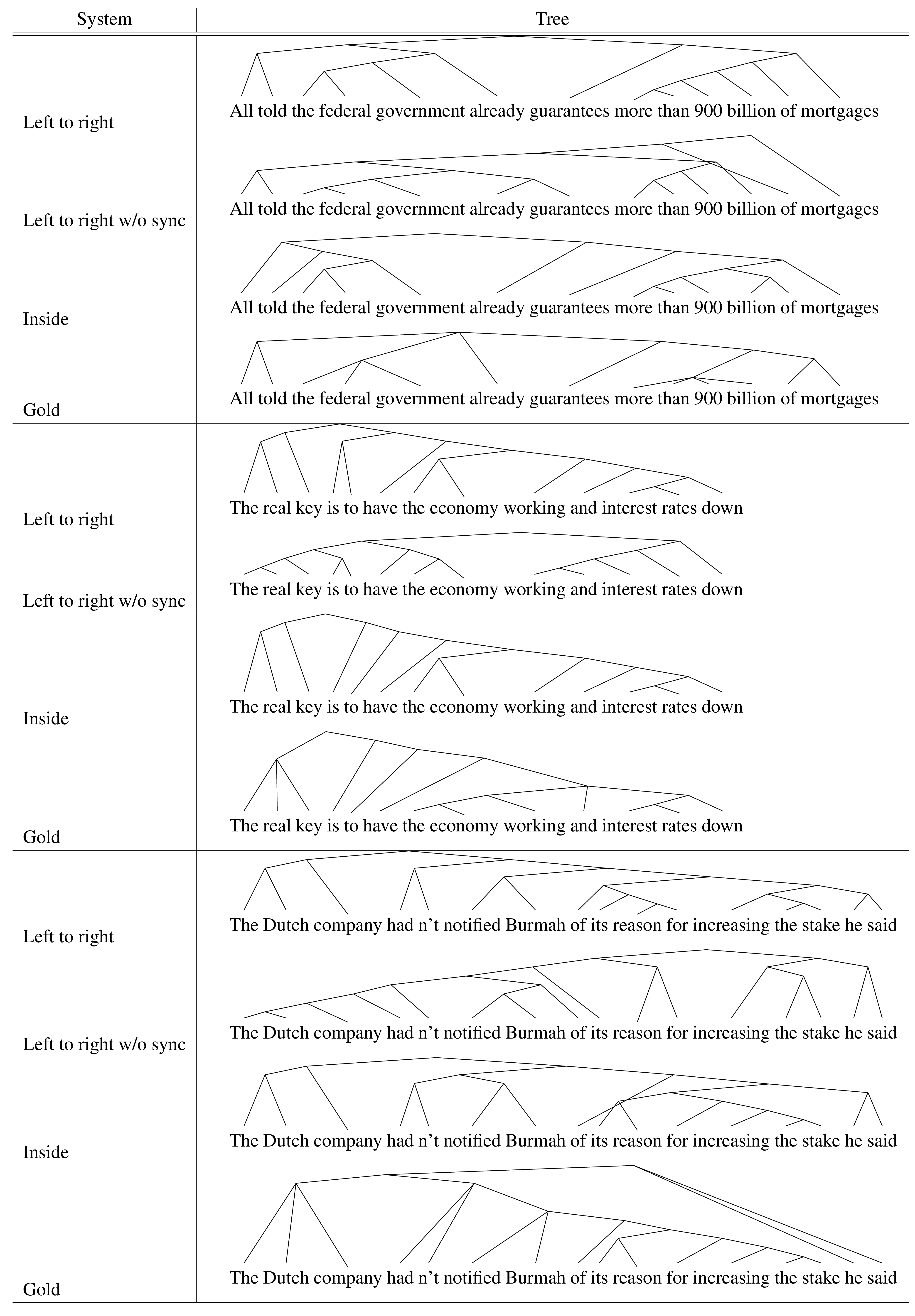}
\end{figure*}